\documentclass{article}

\usepackage[preprint]{neurips_2026}

\usepackage[utf8]{inputenc}
\usepackage[T1]{fontenc}
\usepackage{hyperref}
\usepackage{url}
\usepackage{booktabs}
\usepackage{amsfonts}
\usepackage{amsmath}
\usepackage{amssymb}
\usepackage{nicefrac}
\usepackage{microtype}
\usepackage{xcolor}
\usepackage{graphicx}
\usepackage{subcaption}
\usepackage{multirow}
\usepackage{makecell}
\usepackage{tcolorbox}
\usepackage{tikz}

\title{IntentionNav: A Benchmark for Intent-Driven Object Navigation from Implicit Human Instruction}

\author{
Lin Qian$^{1}$,
Shijie Li$^{2}$,
Sihao Lin$^{3}$,
Xuan Zhang$^{4}$,
Bangya Liu$^{4}$,
Yanran Li$^{5}$,
Hujun Yin$^{1}$\\
$^{1}$The University of Manchester \quad
$^{2}$A*STAR\\
$^{3}$Responsible AI Research Centre, Adelaide University\\
$^{4}$2077AI \quad
$^{5}$University of Bedfordshire
}

\newcommand{\ours}{IntentionNav}
\newcommand{\metric}[1]{\textsc{#1}}

\begin{document}

\maketitle

\begin{abstract}
Existing object navigation benchmarks usually tell an embodied agent which object category to find, such as microwave or chair.
Human-facing embodied AI is often asked something less direct: ``I need something to warm this food'' or ``the room feels stuffy.''
The agent must infer the object that can satisfy the need, find a scene-grounded instance, and decide whether the goal has been reached.
We study this setting as \emph{intent-driven object navigation} and introduce \ours{}, a diagnostic benchmark for active object search from implicit human instructions.
Each episode provides a free-text intent, RGB-D observations, and pose, but withholds the target object name.
\ours{} contains 500 intents over 176 Isaac Sim scenes and 64 target categories.
Each intent is rewritten in four controlled instruction styles and annotated with one of four intent modes, separating surface phrasing from semantic cue type under matched geometry.
This paired design supports analysis of target inference, language robustness, neighborhood reachability, and terminal success rather than only aggregate success.
We evaluated three VLMs using a fixed active-navigation agent.
Models identify the intended target in 48.3\% of episodes and enter its 2 m neighborhood in 68.7\%, but terminate successfully in only 24.9\% and achieve grounded 1 m success in 5.5\%.
Success is highest for event-script intents (28.7\%) and lower for physical-state and affordance intents (19.2\% and 18.5\%), showing that indirect human intent remains a bottleneck for target selection, visual verification, and terminal localization in active embodied search.
\end{abstract}

\section{Introduction}
 
People rarely specify their needs categorically as object labels. A person may say ``I need to rinse these ingredients'' or ``the room feels stuffy'' and expect an agent to infer the implied target from the intent or surrounding scene. This paper explores \emph{intent-driven object navigation}: active object search where the goal is specified only through an implicit human request rather than a category label. This matters for human-centered embodied AI: the difficulty is not only moving through a house, but also grounding a practical, underspecified need in a visible object.

Existing benchmarks cover important pieces of this problem but not the full requirement. Category-goal ObjectNav, REVERIE, and SOON all assume the goal is explicitly provided---as a category label, a referring expression, or an object reference---leaving the agent no ambiguity to resolve \citep{batra2020objectnav,qi2020reverie,zhu2021soon}. Demand-driven navigation (DDN) studies implicit goals, but concentrates on affordance-style demands rather than a paired diagnostic taxonomy over phrasing and semantic cue type \citep{wang2023ddn,mo2024moddn}. Instruction-robustness benchmarks such as ULN vary language explicitness, but target route following rather than object search \citep{nguyen2022uln}. The missing evaluation question is therefore not whether agents can navigate once the goal category is known, but whether they can recover that goal from an indirect human need and carry the inference through active search and terminal localization.

\ours{} formulates an active version of intent-driven object navigation in simulated indoor environments. At each step the agent observes RGB-D and its pose; the target category is never revealed in the instruction. The agent receives no target label: it must resolve the ambiguity of an indirect intent---a single request may be consistent with several plausible object classes---through active exploration, using scene-level visual evidence to ground the goal to a unique physical instance.

A single success rate cannot distinguish whether an agent fails because of how the intent is phrased, what semantic cue it relies on, or where in the navigation chain execution breaks down. \ours{} therefore annotates each intent along two orthogonal axes. \emph{Instruction style} varies the same intent across formal, natural, casual, and emotional phrasing, testing whether the underlying goal survives register shifts. \emph{Intent mode} captures the semantic anchor that implies the target---an event script, an inner state, a physical state of the environment, or a function-only affordance description---testing what kind of inference the agent must perform. Because both axes are applied to the same target set under matched geometry, the benchmark can isolate phrasing effects from cue-type difficulty without constructing separate evaluation sets per condition.

The benchmark is constructed in two stages: automatic candidate generation across 176 scenes produces a pool of 17,624 intent sets, from which 500 items are curated through stratified selection and human validation. Through comprehensive evaluation, we find that agents frequently reach the target neighborhood, overall reachability \metric{OSR} is 0.687 and intent matching \metric{IM} is 0.483, yet terminal success collapses to \metric{SR} 0.249 and grounded success \metric{GSR} to 0.055. The large gap between neighborhood reach and grounded termination reveals that intent understanding alone is an insufficient proxy for embodied task success: failures concentrate in visual evidence gathering and final-state localization after the agent has already arrived near the goal.

Our contributions are threefold. First, we define \emph{intent-driven object navigation} as an active search problem under goal ambiguity, where the agent must recover the target from an indirect human intent and ground it to a unique physical instance through scene-level visual evidence, isolating the target-inference stage that category-goal ObjectNav assumes away. Second, we construct a diagnostic benchmark by curating 500 high-quality intents from a large automatically generated pool of 17,624 candidates across 176 indoor scenes and 64 target categories, yielding 2000 style-controlled instructions with enforced scene-level answerability, physical target uniqueness, and target-name hiding. Third, we introduce paired diagnostic annotations over instruction style and four semantic cue types, enabling stratified analysis of phrasing robustness and intent-cue difficulty on the same target set.
 

\section{Related work}
 
\subsection{Goal specification in embodied navigation}
 
Category-goal ObjectNav established the standard protocol for embodied object search:
the agent receives a target category label and must navigate to any instance
\citep{anderson2018evaluation,batra2020objectnav,chaplot2020semexp,
ramakrishnan2022poni,ramrakhya2022habitatweb}.
Large-scale simulators including Matterport3D, Gibson, Habitat, and HM3D support this
line of work \citep{chang2017matterport,xia2018gibson,savva2019habitat,
szot2021habitat2,ramakrishnan2021hm3d}.
Vision-and-language navigation (VLN) conditions on route instructions
\citep{anderson2018vln,chen2022duet,wang2023scalingvln}, while REVERIE and
SOON use explicit object or region
references that must be grounded visually during navigation
\citep{qi2020reverie,zhu2021soon}.
Recent benchmarks extend the goal space to open-vocabulary objects, multimodal goals,
or semantic tasks \citep{hm3dovon2024,goatbench2024,pin2024,embodiedeval2025}.
Personalized and embodiment-conditioned settings further broaden the scope of
human-facing navigation \citep{ziliotto2025personal,wang2026ucon,su2026capnav}.
Across all these settings the goal is provided explicitly---as a category label, a
route instruction, or a referring expression.
\ours{} changes the goal specification to an implicit human intent: the target
category is hidden, and the agent must recover it from an indirect request before
ordinary navigation mechanisms can engage.
 
\subsection{Intent- and demand-driven navigation}
 
IntentionVG establishes the static grounding baseline for this problem, showing that
grounding difficulty rises sharply when users describe needs rather than naming objects
\citep{wang2024intentionvg}.
\ours{} extends this intent-grounding question to active embodied search, where target
inference, exploration, and terminal localization can fail separately.
DDN and MO-DDN bring this into active navigation: the agent searches for objects
implied by affordance-style demands rather than explicit names
\citep{wang2023ddn,mo2024moddn}.
For example, a demand may ask for something that affords sitting or heating; \ours{}
fixes the hidden target and start pose while varying instruction style and semantic
cue type, so the same physical problem can be compared across language conditions.
CogDDN adds VLM-based dual-process reasoning for demand-driven search
\citep{huang2025cogddn}, while InstructNav and OpenFMNav generalize to free-form
instructions using foundation-model planning and open-vocabulary scene reasoning
\citep{long2024instructnav,kuang2024openfmnav}.
Zero-shot and VLM-based agents such as ZSON, CoWs, VLFM, NavGPT, and NaviLLM further
show that language-model reasoning can decompose instructions, build semantic maps, and
guide exploration under open-vocabulary goals
\citep{majumdar2022zson,gadre2023cows,zhou2023esc,vlfm2024,
zhou2024navgpt,zhou2024navgpt2,chen2024mapgpt,yin2024sgnav,zheng2024navillm}.
These systems differ in whether they rely on learned policies, semantic frontier
selection, scene-graph prompting, or map-guided LLM reasoning, but they still
receive an explicit object or instruction target.
These works collectively show that intent-conditioned search is feasible, but they
optimize for task success rather than diagnostic decomposition: language variation is
a confounder to control, not a variable to measure.
\ours{} fills this gap: rather than proposing a new agent system, it provides the
evaluation infrastructure to measure exactly where any such system breaks down along
the intent-to-navigation chain.
 
\subsection{Diagnostic evaluation of navigation agents}
 
Diagnostic work in VLN has revealed that aggregate success rates can mask qualitatively
different failure modes: systems may collapse under underspecified language, adversarial
phrasing, or behavioral perturbations even when average performance appears robust
\citep{nguyen2022uln,zhu2022diagnosing,fried2023behavioral,liu2021advinstr,
chen2025analogical}.
\ours{} brings this diagnostic philosophy to intent-driven object navigation with two
benchmark-native axes---instruction style and intent mode---that are orthogonal to scene
geometry and target identity.
Because every intent has all four styles and a labelled cue type, the benchmark
supports stratified analysis without constructing separate evaluation sets per condition: it
can reveal whether an agent fails because of phrasing register, semantic cue
difficulty, search coverage, terminal decision-making, or close-range localization
precision.

\section{Benchmark Design}
 
\subsection{Task formulation}

\ours{} defines \emph{intent-driven object navigation} as an active search problem under goal ambiguity. The agent receives an indirect human intent such as ``I need to rinse these ingredients''---a request that is linguistically consistent with multiple object classes and cannot be resolved without visual evidence from the scene. The agent must resolve this ambiguity during exploration: inferring what the intent implies, searching the environment under partial observability, and committing to a final target location, all without access to a target category label at any point during execution. The scene itself is what ultimately disambiguates the goal, making intent understanding and visual grounding inseparable.

\begin{figure}[t]
  \centering
  \includegraphics[width=\linewidth]{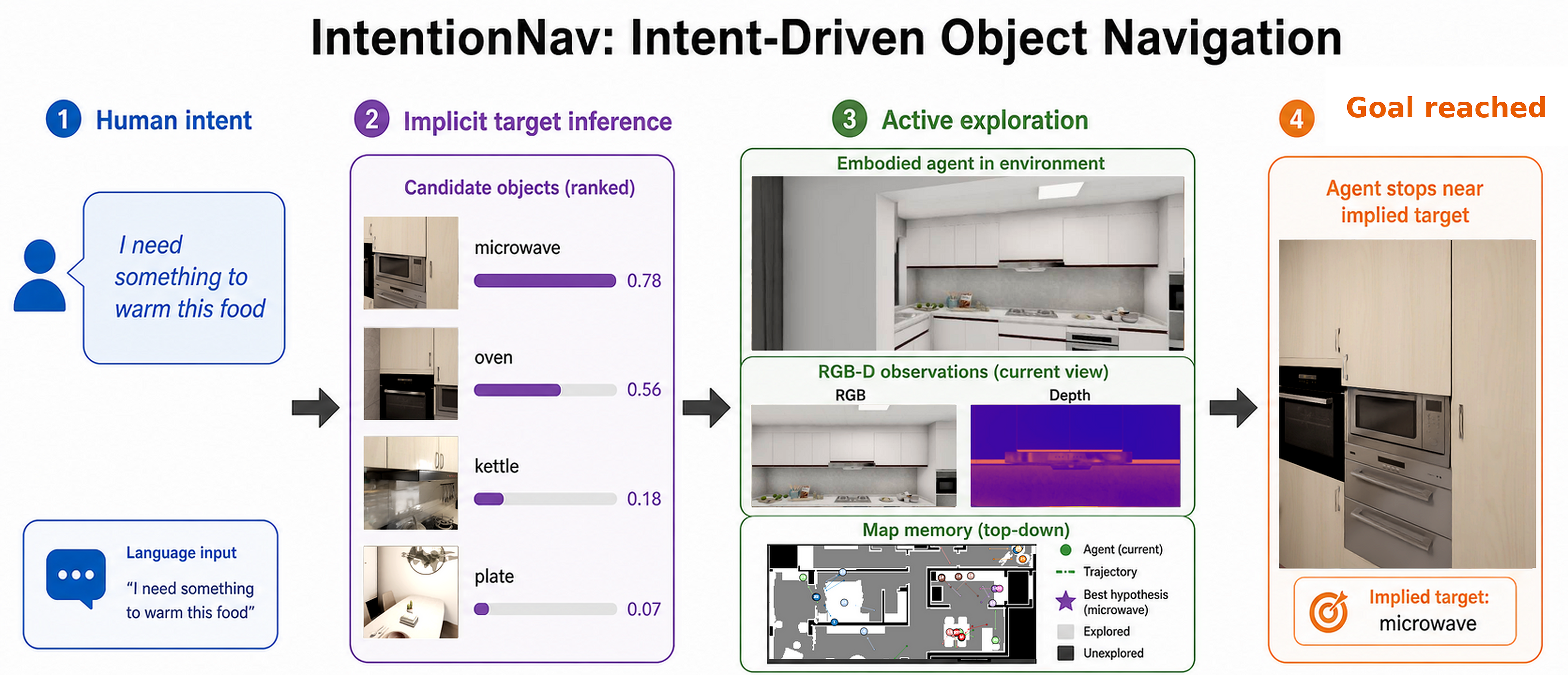}
  \caption{Task overview. \ours{} evaluates intent-driven object navigation: the agent
  receives an indirect human intent, not a target category label, and must infer
  plausible objects, explore with RGB-D and map memory, and take the stop action near
  the scene-grounded target.}
  \label{fig:overview}
\end{figure}

Formally, each scene $s$ exposes a set of physically grounded object candidates
\[
  \mathcal{O}_s=\{(o_i,\,c_i,\,p_i,\,b_i,\,r_i)\}_{i=1}^{N_s},
\]
where $o_i$ is a simulator object identifier, $c_i$ its category, $p_i$ its 3D
position, $b_i$ a bounding volume, and $r_i$ optional local relations (support,
proximity, containment).
An episode is the tuple
\[
  e=(s,\;u,\;q_0,\;\mathcal{O}_s,\;o^\star,\;c^\star,\;p^\star),
\]
where $u$ is the free-text intent, $q_0$ the fixed start pose, and $(o^\star,c^\star,p^\star)$ the hidden ground-truth target. The benchmark enforces two constraints: $u$ must not contain $c^\star$, and $o^\star$ must be the unique admissible instance for $u$ in $s$. Success requires navigating to within a metric radius of $p^\star$ and issuing a stop action; the target category $c^\star$ is retained only for post-hoc evaluation. A task overview is shown in Figure~\ref{fig:overview}.

\subsection{Diagnostic Language Axes }

\ours{} annotates each intent along two orthogonal axes: instruction style and intent
mode.

\textbf{Instruction style} captures how the same intent is expressed across four
register variants: \emph{formal} instructions are explicit and structured;
\emph{natural} instructions reflect everyday conversational phrasing;
\emph{casual} instructions are brief and may omit contextual detail;
\emph{emotional} instructions foreground the speaker's affective state.
Every selected intent has all four style variants generated under target-name hiding,
so style differences in success rate reflect phrasing robustness rather than
variation in scene geometry or target identity.

\textbf{Intent mode} captures the semantic anchor that implies the target.
The four modes are mutually exclusive primaries: \emph{event-script} cues use an
activity or event; \emph{inner-state} cues use the speaker's desire, emotion, or
bodily state; \emph{physical-state} cues use an external environmental condition;
\emph{affordance} cues use only what the target does or looks like.
When multiple anchors are present, the annotation follows the fixed priority
inner-state, event-script, physical-state, then affordance, unless another anchor is
clearly dominant across styles.
On a 100-item double-annotation subset, intent-mode labels achieved 91.0\% raw
agreement and Cohen's $\kappa=0.86$.

\begin{figure}[t]
  \centering
  \includegraphics[width=\linewidth]{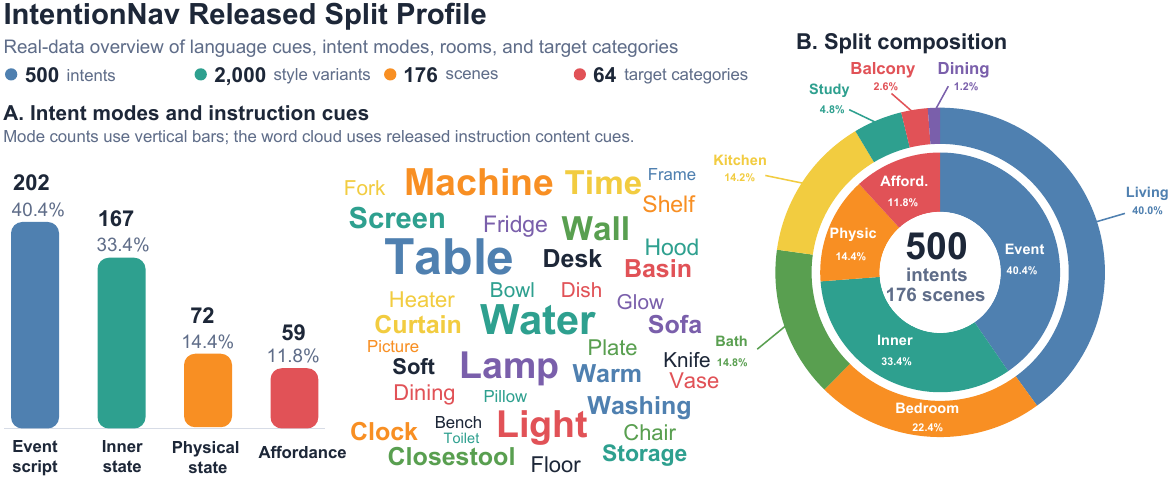}
  \caption{The 4$\times$4 diagnostic language design separates how an intent is phrased
  from what kind of cue grounds the target. The figure summarizes the 500-intent split
  composition, representative language cues, and target-group distribution:
  event-script 202, inner-state 167, physical-state 72, affordance 59; appliances
  133, small objects 115, large furniture 113, lighting/decor 81, and bathroom
  objects 58.}
  \label{fig:taxonomy}
\end{figure}

The two axes answer different scientific questions.
Style asks whether the same underlying goal survives register shifts.
Mode asks what kind of inference is needed: event-script and inner-state cues can
invoke common routines or needs, whereas affordance cues require mapping a function
to possible objects without naming the object.
Together they support stratified analysis on the same target set without constructing
separate evaluation sets per condition.
Figure~\ref{fig:taxonomy} illustrates both axes and the resulting benchmark
composition.

\subsection{Benchmark construction}

\ours{} is constructed in two stages: automatic candidate generation followed by
human-validated curation.

In the first stage, we enumerate support surfaces, candidate targets, canonical
camera views, and local relations from the simulator scene graph and rendered
observations across 176 indoor scenes drawn from the VLNVerse simulation
environment \citep{lin2025vlnverse}, yielding 3,871 surface records and 4,406
view-target observations.
A language annotation step then generates four candidate intent sets per
observation, each paired with four controlled English styles, producing a
candidate pool of \textbf{17,624 intent sets}.
This stage prioritizes coverage: the pool spans diverse scene types, target
categories, and intent phrasings without imposing balance or quality constraints.

In the second stage, we curate the final \textbf{500-intent diagnostic split} through stratified selection and human validation. Selection jointly optimizes category balance, scene diversity, visual answerability, and physical target uniqueness; items with physically ambiguous targets are excluded to ensure the benchmark tests intent grounding rather than arbitrary tie-breaking. A stratified 250-item sample (50\% of base intents, 1000 style-specific instructions) was manually audited, checking target-name leakage, scene answerability, physical uniqueness, style preservation, and simulator executability, reaching 100\% pass rates after correction. The final consistency audit replaced 17 non-unique and 10 visually invalid selections using held-out candidates, preserving the 500-item size. The 35:1 pool-to-final ratio reflects the strictness of the quality gate; the curated split keeps evaluation cost manageable across model families while the balanced category and scene distribution ensures that aggregate scores reflect performance across the full intended scope of the task.

The 500 selected items span 176 scenes and 64 target categories, with all four English styles yielding 2000 instructions. Intent-type distribution: event-script 202, inner-state 167, physical-state 72, affordance 59. Target groups: appliances 133, small objects 115, large furniture 113, lighting/decor 81, bathroom objects 58. Room coverage is concentrated in living rooms (200), bedrooms (112), bathrooms (74), and kitchens (71); see Table~\ref{tab:object-groups} in the appendix
for the full category breakdown.

Episode starts are sampled with deterministic seeds and target distance constraints,
so that every style variant of the same intent is evaluated from matched geometry.
This paired design ensures that style differences in success rate reflect phrasing
effects, not confounds from different start poses or target placements.
 
\textbf{Validation and Consistency Audit.}
Because intents are generated automatically, the raw pool may contain target-name leakage or stylistically uniform phrasing that would allow an agent to recover the target by lexical matching rather than genuine intent understanding. Items are therefore validated along three dimensions before inclusion. Lexical leakage control verifies that no style variant names or closely synonymizes the target category. Language refinement confirms indirect evidence and style separation, requiring that at least two variants rely on contextual or affordance-based cues rather than direct description. Independent visual-language scoring rates target recoverability and style distinguishability on 0--10 scales; items below threshold are revised for up to three rounds or replaced. 

\begin{figure}[t]
  \centering
  \begin{minipage}[t]{0.40\linewidth}
    \vspace{0pt}
    \includegraphics[width=\linewidth,trim={0 0 300 0},clip]{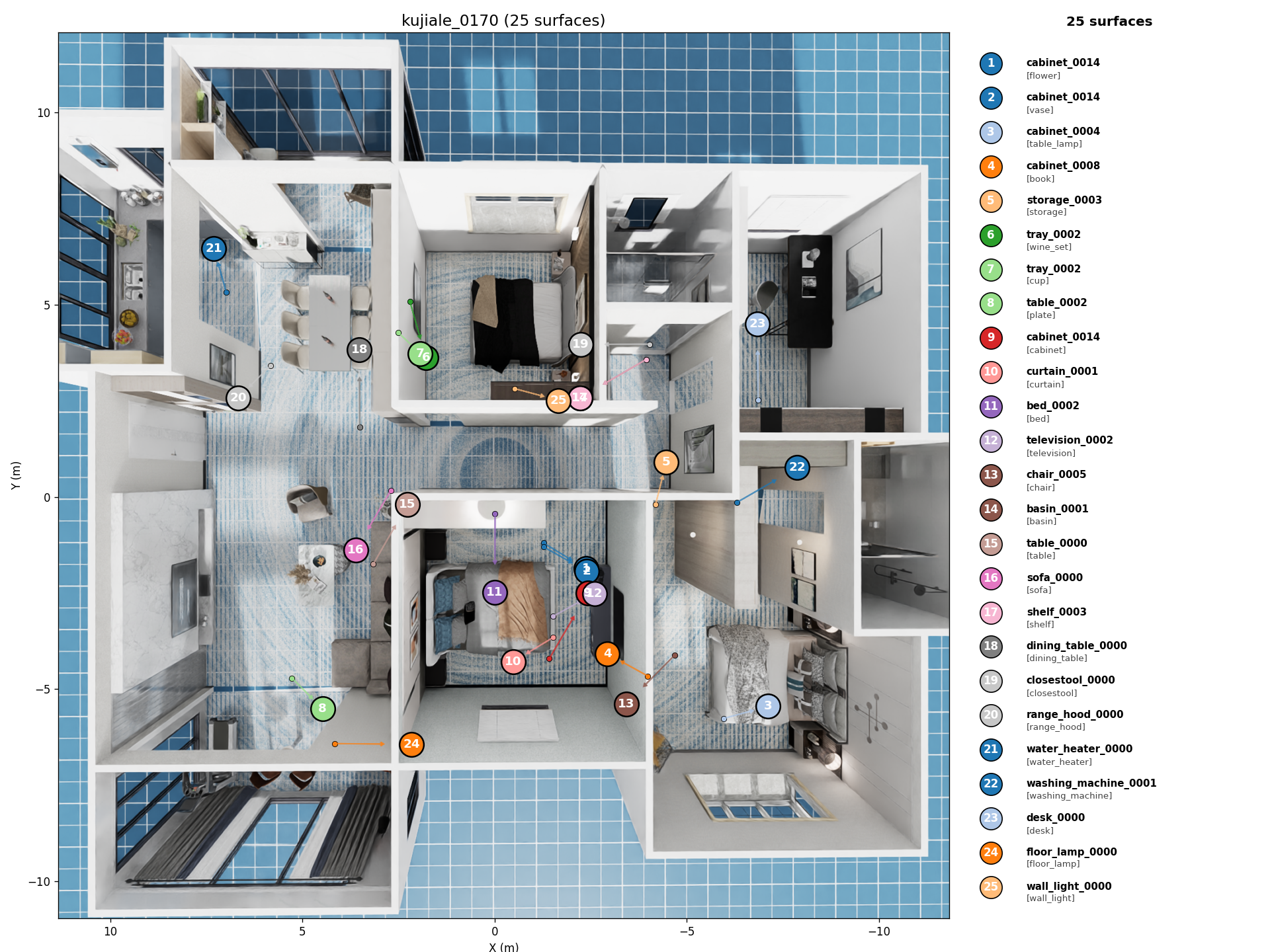}
  \end{minipage}\hspace{0.006\linewidth}%
  \begin{minipage}[t]{0.592\linewidth}
    \vspace{0pt}
    \centering
    \vspace{4pt}
    \begin{minipage}{0.333\linewidth}
      \centering
      \includegraphics[width=\linewidth]{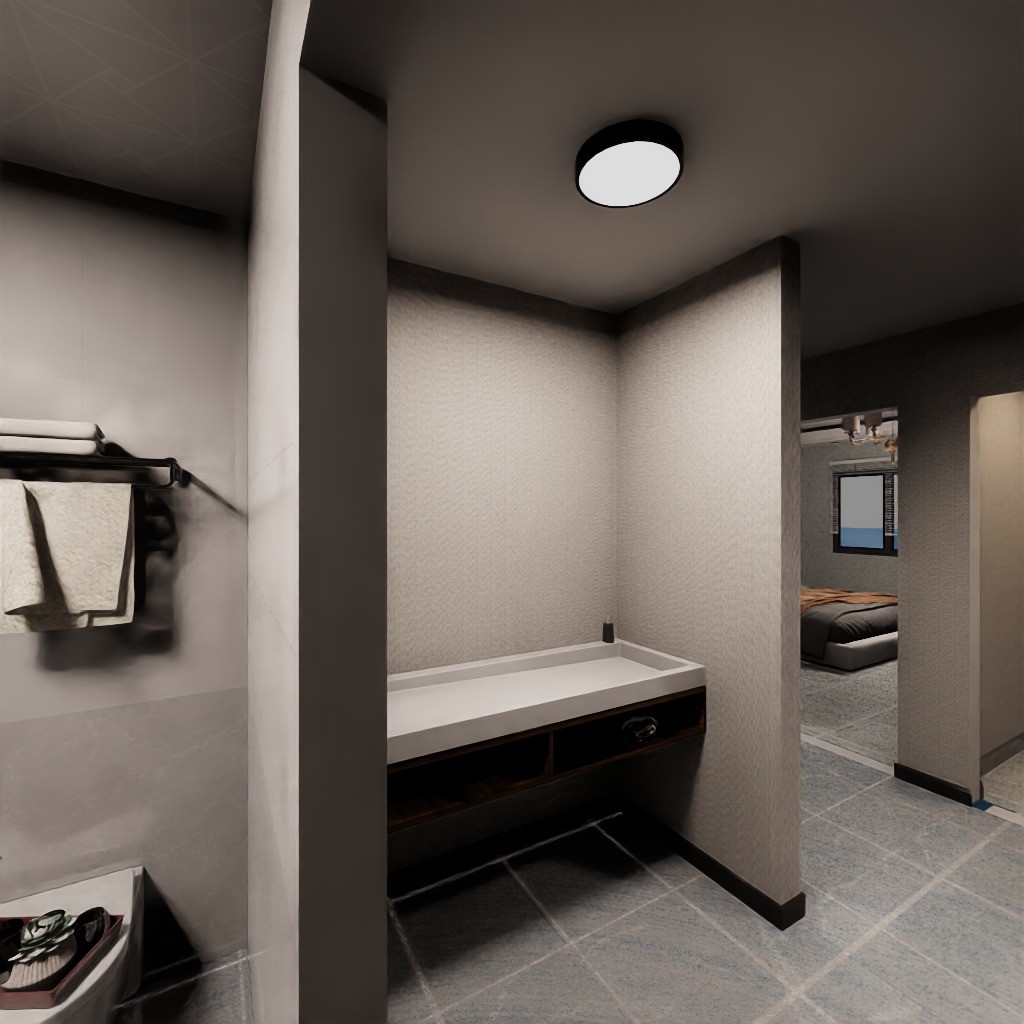}\\[-1pt]
      \scriptsize Basin\\living area
    \end{minipage}%
    \begin{minipage}{0.333\linewidth}
      \centering
      \includegraphics[width=\linewidth]{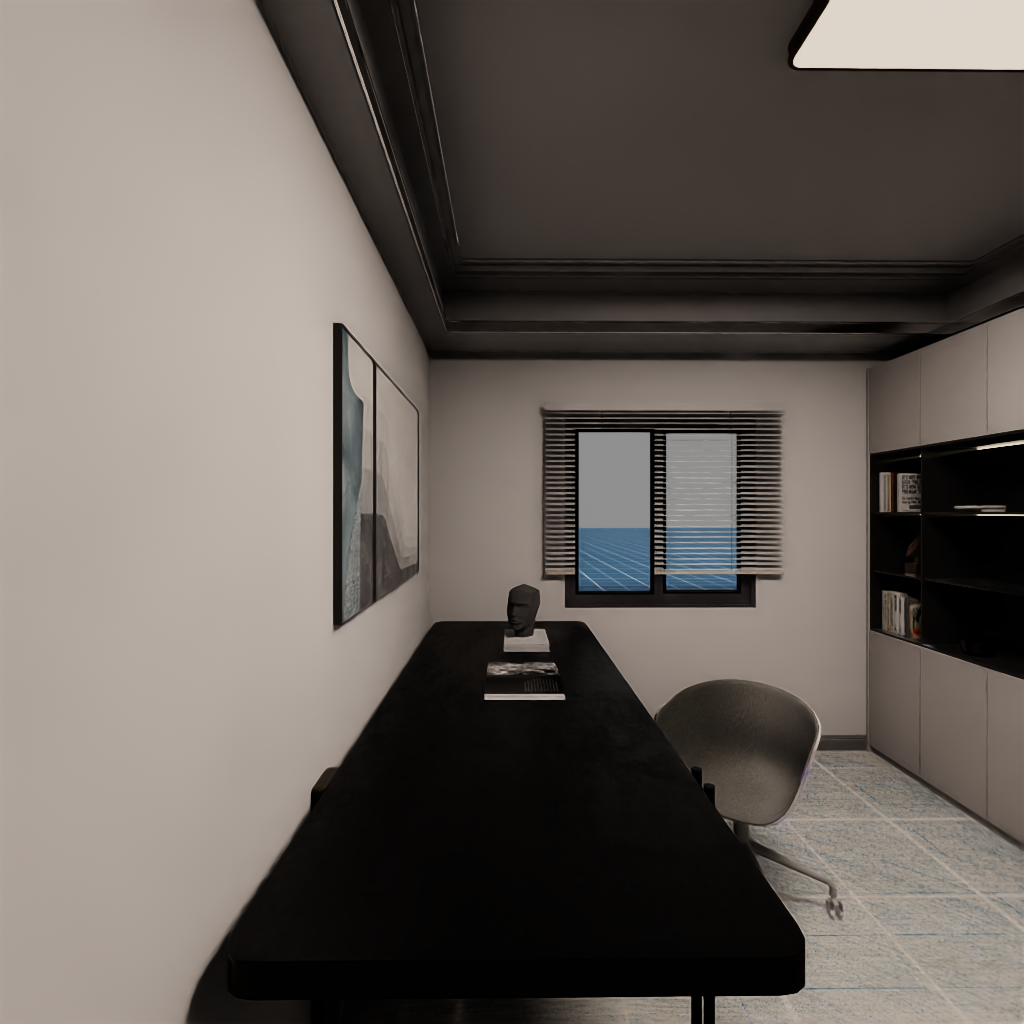}\\[-1pt]
      \scriptsize Desk\\study room
    \end{minipage}%
    \begin{minipage}{0.333\linewidth}
      \centering
      \includegraphics[width=\linewidth]{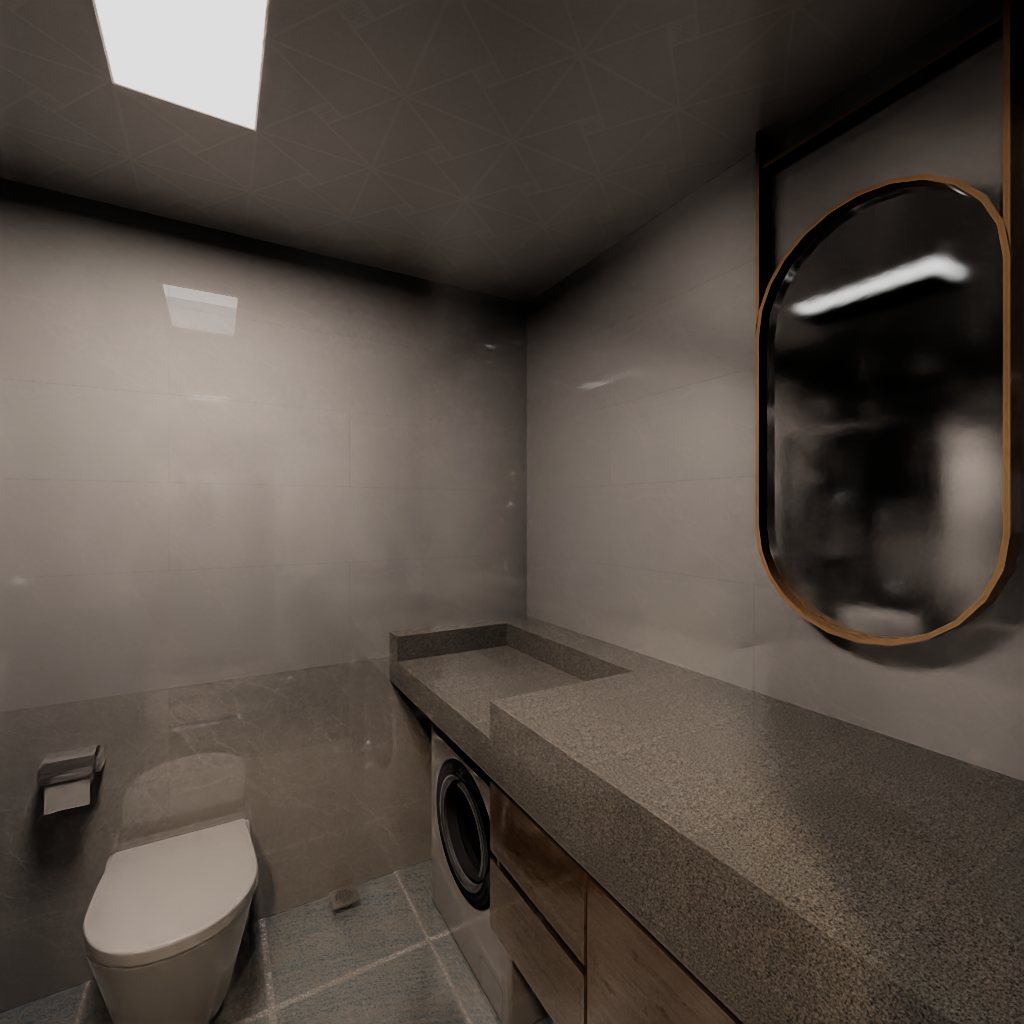}\\[-1pt]
      \scriptsize Washer\\bathroom
    \end{minipage}
    \vspace{1pt}
    \scriptsize
    \setlength{\tabcolsep}{2pt}
    \begin{tabular}{@{}lll@{}}
      \toprule
      Target & Room cue & Graph cue \\
      \midrule
      Basin & living area & sink-like fixture \\
      Desk & study room & work surface \\
      Washer & bathroom & appliance node \\
      \bottomrule
    \end{tabular}
  \end{minipage}
  \caption{Candidate acquisition before benchmark filtering. Left: a rendered scene
  with 25 scene-graph-derived target viewpoints, where each numbered circle denotes
  one sampled target-object view. Right: three target-view crops from the same scene.
  The raw pool is filtered for physical uniqueness, lexical leakage, intent
  recoverability, and style separation before constructing the 500-intent split.}
  \label{fig:kujiale-example}
\end{figure}

\subsection{Comparison with prior benchmarks}
 
Table~\ref{tab:benchmark-comparison} positions \ours{} against nearby navigation benchmarks. The key distinction is not scale alone, but the combination of implicit target inference, scene-level disambiguation, style variation, cue-type taxonomy, and stratified evaluation. Benchmarks such as GOAT-Bench, PIN, PersONAL, and UcON broaden what counts as a goal; systems such as OpenFMNav and InstructNav broaden what instructions an agent can follow. \ours{} makes the language variation itself the object of measurement.
 
\begin{table}[h]
  \caption{Comparison with prior embodied navigation benchmarks. ``Partial'' indicates
  that the property is present but not a primary controlled axis.}
  \label{tab:benchmark-comparison}
  \centering
  \small
  \setlength{\tabcolsep}{4pt}
  \renewcommand{\arraystretch}{1.15}
  \begin{tabular}{lccccc}
    \toprule
    Benchmark & Implicit & Scene disambig. & Multi-style & Multi-cue & Diagnostic \\
    \midrule
    ObjectNav \citep{batra2020objectnav} & \texttimes & \texttimes & \texttimes & \texttimes & \texttimes \\
    REVERIE \citep{qi2020reverie}        & \texttimes & \checkmark & \texttimes & \texttimes & $\circ$ \\
    SOON \citep{zhu2021soon}             & \texttimes & \checkmark & $\circ$    & \texttimes & $\circ$ \\
    DDN/MO-DDN \citep{wang2023ddn,mo2024moddn} & \checkmark & $\circ$ & \texttimes & $\circ$ & \texttimes \\
    ULN \citep{nguyen2022uln}            & \texttimes & \texttimes & \checkmark & \texttimes & \checkmark \\
    CoWs \citep{gadre2023cows}           & \texttimes & \texttimes & \texttimes & $\circ$    & \texttimes \\
    \midrule
    \textbf{\ours{}}                     & \checkmark & \checkmark & \checkmark & \checkmark & \checkmark \\
    \bottomrule
  \end{tabular}
  \vspace{0.3em}
  
  \footnotesize\raggedright
  \checkmark\ = fully supported;\quad $\circ$\ = partial;\quad \texttimes\ = not supported
\end{table}

\section{Evaluation Framework}

Evaluating intent-driven object navigation requires more than a single success rate. An agent may correctly infer the target category yet fail to locate the instance; it may reach the target neighborhood but finish outside it; it may end at the right location yet without visual confirmation of the target. A single aggregate metric collapses these qualitatively different failure modes and obscures where improvement is needed. \ours{} therefore structures evaluation around the intent-to-navigation chain, assigning a dedicated metric to each stage: intent inference (\metric{IM}), neighborhood reachability (\metric{OSR}), terminal success (\metric{SR}), and grounded localization (\metric{GSR}). The diagnostic gaps between consecutive stages quantify how much is lost at each transition and direct attention to the dominant bottleneck.

\subsection{Reference agent}

We evaluated all models with the same reference active-navigation agent. At each step, the agent receives RGB-D observations from Isaac Sim-based
VLNVerse environment \citep{lin2025vlnverse}, uses the VLM to estimate target visibility and exploration direction, applies open-vocabulary object detection \citep{liu2023groundingdino}, maintains target evidence, and plans toward an informative waypoint on the navigable map. An episode ends when visual evidence and geometric proximity support the termination condition, or after a fixed 30-step budget; this doubles the roughly 15 human waypoint decisions observed in manual route inspection, and 40-step pilot checks mostly add low-efficiency wandering without changing the main \metric{SR}/\metric{OSR} ordering. The agent is modular: future systems can replace target inference, perception, memory, planning, or the termination criterion while preserving the same 500-intent evaluation split. We call this system the \emph{Reference Agent} rather than a baseline to avoid overclaiming a method contribution; its role is to hold all non-VLM components fixed so that VLM-swap experiments measure the full intent-to-navigation chain.

\subsection{Evaluation metrics}
\label{sec:evaluation-metrics}

The metric suite assesses target inference, neighborhood reachability, grounded
termination, and path efficiency.

\textbf{Intent matching} measures whether the agent identifies the correct target
category:
\[
  \operatorname{IM}(e)=\mathbf{1}[\operatorname{syn}(\hat{c})=\operatorname{syn}(c^\star)],
\]
where $\hat{c}$ is the predicted category and $\operatorname{syn}(\cdot)$ maps strings to the submitted synonym vocabulary (e.g., sofa/couch, refrigerator/fridge).
\metric{IM} measures category recovery rather than exact wording; because the
vocabulary is finite, it should be read alongside navigation metrics rather than as a
standalone measure of intent understanding.
The synonym vocabulary and construction judge prompt are detailed in
Appendix~\ref{app:synonyms-prompts}.

\textbf{Navigation success} is measured at a headline radius $r$:
\[
  \operatorname{SR}_{r}(e)=\mathbf{1}[d_\tau\le r],
  \qquad
  \operatorname{OSR}_{r}(e)=\mathbf{1}\!\left[\min_{t\le \tau} d_t\le r\right],
\]
where $d_t=\lVert q_t-p^\star\rVert_2$ is the distance to the target position
$p^\star$ at step $t$, and $\tau$ is the terminal step.
\metric{SR} requires the terminal position to be within radius $r$; \metric{OSR}
records same-trajectory reachability---first entry into the neighborhood counts even
if the final position is outside the radius.

\textbf{Grounded success and path efficiency}:
\[
  \operatorname{GSR}_{r}(e)=\operatorname{SR}_{r}(e)\,\mathbf{1}[v_\tau=1],
  \qquad
  \operatorname{SPL}_{r}(e)=\operatorname{SR}_{r}(e)\frac{L^\star}{\max(L,L^\star)},
\]
where $v_\tau$ is final-frame target visibility and $L^\star$, $L$ are shortest-path
and actual trajectory lengths.
\metric{GSR} additionally requires the target to be visible at the terminal frame,
enforcing close-range visual confirmation beyond geometric proximity.

All navigation metrics use $r=2.0$\,m except \metric{GSR}, which uses the stricter
$r=1.0$\,m.
\metric{TL} is mean trajectory length over successful \metric{SR}@2m episodes;
unlike \metric{SPL}, it is conditional on success and does not penalize failures.
\metric{CSR} is cross-style robustness: the fraction of intents for which all four
style variants succeed for the same model.

Split-level scores average over the evaluated split $\mathcal{D}$:
\[
  \overline{M}=\frac{1}{|\mathcal{D}|}\sum_{e\in\mathcal{D}} M(e).
\]

We additionally report three diagnostic gaps that localize failures along the
intent-to-navigation chain:
\[
  g_{\text{infer}}=\overline{\operatorname{IM}}-\overline{\operatorname{SR}_{2.0}},
  \quad
  g_{\text{term}}=\overline{\operatorname{OSR}_{2.0}}-\overline{\operatorname{SR}_{2.0}},
  \quad
  g_{\text{prec}}=\overline{\operatorname{SR}_{2.0}}-\overline{\operatorname{GSR}_{1.0}}.
\]
For robustness analysis, $\Delta_{\text{style}}$ and $\Delta_{\text{intent}}$ are
max-minus-min \metric{SR} across instruction styles and intent modes.

\section{Experiments}
\label{sec:experiments}
 
We evaluated three VLMs: GPT-5.4, Qwen3.6-Plus, and Gemini-3.1-Flash-Lite, on the full 500-intent split under all four instruction styles, yielding 2000 zero-shot episodes per model. All models shared the same reference agent, prompts, detector, map construction, and termination criterion; only the VLM backend varies. This design treated the three models as controlled perturbations of a fixed system rather than a provider ranking: differences in aggregate \metric{SR} reflect sensitivity of the full intent-to-navigation chain to the VLM component, not an endorsement of any particular model family.
 
\subsection{Main results}

Table~\ref{tab:main-results} reveals a consistent pattern across all three models:
intent inference and neighborhood reachability are substantially easier than terminal
success.
Overall \metric{IM} is 0.483 and \metric{OSR} is 0.687, yet \metric{SR} drops to
0.249 and \metric{GSR} to 0.055.
The three VLMs differ by only 1.6 percentage points in \metric{SR}, so the model
ordering is descriptive rather than statistically meaningful---the dominant signal is
the shared gap structure, not inter-model differences.
The intent gap ($g_{\text{infer}}=0.234$) and the reach-to-success gap
($g_{\text{term}}=0.438$) both exceed the spread among VLMs, confirming that the
benchmark measures a system-level chain rather than ranking model families.
Figure~\ref{fig:diagnostic-radar} in the appendix summarizes the diagnostic
profile of all three models across the full metric suite.

Among models, Gemini-3.1-Flash-Lite achieves the highest \metric{SR}, \metric{GSR},
and \metric{SPL} with the shortest successful trajectories (22.7\,m), while GPT-5.4
ties Gemini on \metric{IM} and narrowly leads on \metric{OSR}.
The divergence between \metric{IM} and \metric{SR} across models illustrates why
no single metric suffices: a model that leads on intent inference does not
necessarily lead on terminal success, and neighborhood reachability can be high
while grounded success remains low.

\begin{table}[h]
  \caption{Main full-benchmark results over 2000 episodes per model. \metric{SR}/\metric{OSR}/%
  \metric{SPL} use 2.0\,m; \metric{GSR} uses 1.0\,m plus final-frame visibility.
  \metric{TL} is successful path length in meters (lower is better).}
  \label{tab:main-results}
  \centering
  \footnotesize
  \setlength{\tabcolsep}{3.6pt}
  \begin{tabular}{lcccccc}
    \toprule
    Model & \metric{IM} & \metric{SR} & \metric{GSR} & \metric{OSR} & \metric{SPL} & \metric{TL} \\
    \midrule
    GPT-5.4                  & \textbf{0.504} & 0.249          & 0.051          & \textbf{0.691} & 0.078          & 24.3 \\
    Qwen3.6-Plus             & 0.441          & 0.241          & 0.048          & 0.681          & 0.072          & 25.1 \\
    Gemini-3.1-Flash-Lite    & \textbf{0.504} & \textbf{0.257} & \textbf{0.066} & 0.690          & \textbf{0.087} & \textbf{22.7} \\
    \midrule
    Overall                  & 0.483          & 0.249          & 0.055          & 0.687          & 0.079          & 24.0 \\
    \bottomrule
  \end{tabular}
\end{table}

\subsection{Explicit-target diagnostic}

To isolate the contribution of intent inference to overall failure, we run a
controlled ablation on the Gemini formal setting in which the implicit intent is
replaced by the ground-truth target category, while scenes, starts, perception,
mapping, and the termination criterion are held fixed.
Naming the target removes the intent-inference step entirely, so \metric{IM} is not
reported for this condition.

The explicit-target setting reaches \metric{SR}@2m / \metric{GSR}@1m / \metric{OSR}@2m
of 0.448 / 0.124 / 0.776 (\metric{SR}@2m 95\% CI 0.408--0.494), compared with
0.280 / 0.066 / 0.706 for the implicit Gemini formal condition.
The 16.8-point \metric{SR} gain confirms that intent inference is a genuine bottleneck.
However, explicit-target \metric{SR} of 0.448 still leaves more than half of episodes
failing, and \metric{GSR} reaches only 0.124---demonstrating that visual evidence
gathering, last-meter localization, and final-state precision remain major bottlenecks
even after the target category is provided.

\subsection{Diagnostic analysis}

\textbf{Instruction style.}
Table~\ref{tab:style} shows that style sensitivity is modest in aggregate: mean
\metric{SR} ranges from 0.239 (casual) to 0.267 (formal), a spread of 2.8 points.
Because style variants share the same scene, target, and start distribution, this gap
reflects phrasing effects alone.
At the item level, however, \metric{CSR} tells a different story: only 5.6--9.6\% of
intents succeed under all four styles for the same model, showing that aggregate
robustness can coexist with substantial per-intent inconsistency.

\begin{table}[h]
  \caption{Success rate by model and instruction style. Per-style scores are over
  500 episodes. $\Delta_{\text{style}}$ is max minus min \metric{SR}; \metric{CSR} is
  the fraction of selected intents where all four styles succeed.}
  \label{tab:style}
  \centering
  \small
  \begin{tabular}{lcccccc}
    \toprule
    Model & Formal & Natural & Casual & Emotional & $\Delta_{\text{style}}$ & \metric{CSR} \\
    \midrule
    GPT-5.4               & 0.268 & 0.246 & 0.244 & 0.240 & 0.028 & 0.056 \\
    Qwen3.6-Plus          & 0.254 & 0.236 & 0.222 & 0.254 & 0.032 & 0.070 \\
    Gemini-3.1-Flash-Lite & 0.280 & 0.246 & 0.250 & 0.252 & 0.034 & 0.096 \\
    \midrule
    Mean & \textbf{0.267} & \textbf{0.243} & \textbf{0.239} & \textbf{0.249} & -- & -- \\
    \bottomrule
  \end{tabular}
\end{table}

\textbf{Intent mode.}
Table~\ref{tab:intent-mode} shows that intent mode is the stronger diagnostic split.
Event-script intents reach 0.287 mean \metric{SR}, while physical-state and affordance
intents reach only 0.192 and 0.185---a gap of over 10 points driven by the type of
inference required rather than phrasing.
Affordance intents show the highest \metric{OSR} (0.809) but low terminal success,
indicating that agents enter plausible neighborhoods without obtaining sufficient
visual evidence to stop confidently.
Physical-state intents show lower \metric{OSR} (0.567), pointing to a search
bottleneck: when the instruction describes an environmental condition, agents struggle
to identify which object to look for and where.

\begin{table}[h]
  \caption{Success rate by intent mode. Per-mode scores are averaged over episodes in
  that mode. $\Delta_{\text{intent}}$ is max minus min \metric{SR}.}
  \label{tab:intent-mode}
  \centering
  \small
  \begin{tabular}{lccccc}
    \toprule
    Model & Event & Inner & Physical & Afford. & $\Delta_{\text{intent}}$ \\
    \midrule
    GPT-5.4               & 0.280 & 0.260 & 0.208 & 0.165 & 0.114 \\
    Qwen3.6-Plus          & 0.287 & 0.225 & 0.181 & 0.208 & 0.107 \\
    Gemini-3.1-Flash-Lite & 0.295 & 0.268 & 0.188 & 0.182 & 0.112 \\
    \midrule
    Mean & \textbf{0.287} & \textbf{0.251} & \textbf{0.192} & \textbf{0.185} & -- \\
    \bottomrule
  \end{tabular}
\end{table}

\textbf{Stage-wise gap decomposition.}
Table~\ref{tab:gap-decomposition} decomposes failures along the intent-to-navigation
chain.
The reach-to-success gap ($g_{\text{term}}=0.438$) is the dominant bottleneck,
nearly twice the intent gap ($g_{\text{infer}}=0.234$): agents more often fail after
reaching the target neighborhood than before finding it.
The precision gap ($g_{\text{prec}}=0.194$) further shows that even successful final positions
frequently lack visual confirmation.
Radius sensitivity reinforces this: geometric \metric{SR} rises from 0.120 at 1\,m
to 0.346 at 3\,m, while grounded \metric{GSR}@1m remains 0.055 (full radius breakdown
in Table~\ref{tab:radius-sensitivity}).

\begin{table}[h]
  \caption{Diagnostic gaps along the intent-to-navigation chain. Larger values indicate greater loss between consecutive stages.}
  \label{tab:gap-decomposition}
  \centering
  \footnotesize
  \setlength{\tabcolsep}{3.2pt}
  \begin{tabular}{lccccc}
    \toprule
    Model 
    & \metric{IM}$-$\metric{SR} 
    & \metric{OSR}$-$\metric{SR} 
    & \metric{SR}$-$\metric{GSR} 
    & $\Delta_{\text{style}}$ 
    & $\Delta_{\text{intent}}$ \\
    \midrule
    GPT-5.4               
    & 0.255 & 0.441 & 0.199 & 0.028 & 0.114 \\
    Qwen3.6-Plus          
    & 0.199 & 0.440 & 0.194 & 0.032 & 0.107 \\
    Gemini-3.1-Flash-Lite 
    & 0.247 & 0.433 & 0.191 & 0.034 & 0.112 \\
    \midrule
    Overall               
    & \textbf{0.234} 
    & \textbf{0.438} 
    & \textbf{0.194}
    & \textbf{0.031} 
    & \textbf{0.111} \\
    \bottomrule
  \end{tabular}
\end{table}

Failure logs reinforce the final-state bottleneck: crediting the first 2.0\,m
neighborhood entry as success would raise \metric{SR} from 0.249 to 0.687,
matching \metric{OSR} exactly.
Qualitative inspection reveals unstable affordance hypotheses, reached-but-unverified
objects, and weak target-verification evidence even when the agent has entered the
target neighborhood.
These patterns share a common cause: the agent's final state depends on
instantaneous visual evidence rather than accumulated confidence, making it
sensitive to viewpoint and occlusion at close range.
\section{Conclusions}
 
\ours{} formalizes intent-driven ObjectNav as a diagnostic benchmark for evaluating
whether embodied agents can infer scene-grounded target categories from indirect
human intent and realize those in active search.
The benchmark couples 500 physically grounded targets with four matched instruction
styles, semantic-cue labels, leakage controls, target-uniqueness filtering, and fixed
evaluation starts, yielding 2000 controlled instructions.
Across three VLMs, each evaluated on the same 2000-episode split, models often infer plausible targets and enter the target
neighborhood, yet fail to terminate successfully: \metric{IM} 0.483 and
\metric{OSR} 0.687 contrast with \metric{SR} 0.249 and \metric{GSR} 0.055.
The explicit-target diagnostic shows that naming the object improves success but does
not eliminate bottlenecks in visual evidence gathering, close-range localization, and
terminal decision-making.
By fixing physical targets while varying linguistic form and intent type, \ours{}
provides a controlled basis for diagnosing how VLM-based embodied agents handle more
realistic human-facing navigation requests.  \ours{} is intentionally scoped to
controlled simulation: English single-target instructions, static apartments, no
clarification dialogue, a 30-step horizon, and one reference agent using
open-vocabulary detection and hosted VLMs.
Multilingual, multi-object, interactive, and real-home deployments remain orthogonal
extensions; finite synonym matching and automated rewriting and judging also bound
diagnostic precision.
The results should therefore be read as benchmark evidence about intent-to-navigation
bottlenecks, not as a claim of deployment readiness.
 
\newpage
\bibliographystyle{plain}
\bibliography{references}

\appendix

\section{Additional benchmark statistics}

Table~\ref{tab:object-groups} lists the five target groups and representative
categories in the 500-intent split.
Table~\ref{tab:radius-sensitivity} reports geometric and grounded success at three
success radii, complementing the headline 2.0\,m results in the main paper.

\begin{table}[h]
  \caption{Object-group distribution in the 500-intent benchmark.}
  \label{tab:object-groups}
  \centering
  \small
  \begin{tabular}{lcl}
    \toprule
    Group & Count & Example categories \\
    \midrule
    Appliances     & 133 & fridge, built-in oven, range hood, kettle \\
    Small objects  & 115 & book, cup, plate, fork, spoon, vase \\
    Large furniture & 113 & sofa, bed, dining table, desk, shelf \\
    Lighting/decor &  81 & chandelier, table lamp, sculpture, menorah \\
    Bathroom       &  58 & bathtub, basin, mirror, toilet, curtain \\
    \bottomrule
  \end{tabular}
\end{table}

\begin{table}[h]
  \caption{Radius sensitivity on the shared 2000-episode split, reported per model and
  averaged across three models. \metric{SR} is
  geometric success at the listed radius; \metric{GSR} additionally requires
  final-frame target visibility at 1.0\,m.}
  \label{tab:radius-sensitivity}
  \centering
  \small
  \begin{tabular}{lcccc}
    \toprule
    Model & \metric{SR}@1m & \metric{SR}@2m & \metric{SR}@3m & \metric{GSR}@1m \\
    \midrule
    GPT-5.4               & 0.120 & 0.249 & 0.347 & 0.051 \\
    Qwen3.6-Plus          & 0.113 & 0.241 & 0.341 & 0.048 \\
    Gemini-3.1-Flash-Lite & 0.128 & 0.257 & 0.351 & 0.066 \\
    \midrule
    Overall               & \textbf{0.120} & \textbf{0.249} & \textbf{0.346}
                          & \textbf{0.055} \\
    \bottomrule
  \end{tabular}
\end{table}

\section{Reproducibility notes}

The reported evaluation uses the selected 500-intent split defined in the paper.
All reported results use the reference agent with fixed prompts, detector
settings, and metric definitions; metrics are computed by the official evaluation
implementation with synonym-aware intent matching from the benchmark vocabulary.

Reported runs were conducted on local NVIDIA GPU workstations.
The complete three-model evaluation over the shared 2000-episode split required approximately 150 GPU-hours on an RTX 5090;
wall-clock time varies because externally hosted VLM API latency introduces scheduling
variance.

The submitted artifact includes the reference-agent implementation, prompt and
configuration files, synonym vocabulary, aggregation scripts, and table-generation
utilities. It does not include the full benchmark data assets, scene renders, episode logs,
or simulator-derived scene packages; the scripts can regenerate the reported tables
when provided with compatible raw episode logs.
The paper identifies the main upstream dependencies, including Isaac Sim scene
assets, Grounding-DINO, and external VLM services.

\section{Reference navigation pipeline}
\label{app:navigation-pipeline}

\begin{figure}[h]
  \centering
  \includegraphics[width=\linewidth]{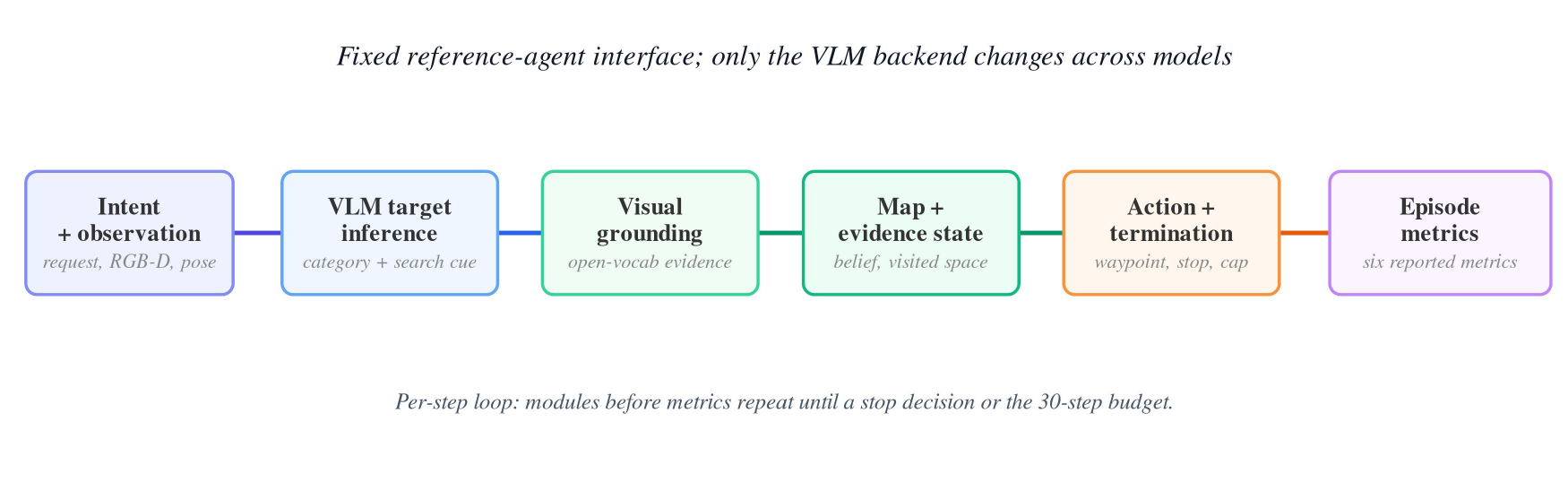}
  \caption{Reference navigation pipeline used for all evaluated VLMs. The agent
  repeatedly combines the implicit request with the current observation, infers a
  target category, updates visual evidence and map state, and selects the next
  waypoint or termination action. The same interface, step budget, and metrics are
  used across model backends.}
  \label{fig:navigation-pipeline}
\end{figure}

Figure~\ref{fig:navigation-pipeline} summarizes the fixed reference-agent pipeline
used for all evaluated VLM backends.
The diagram is intended to clarify the evaluation interface rather than introduce a
new navigation algorithm: the VLM component is swapped across models, while
perception, evidence memory, planning, termination, and metric aggregation are held
fixed.

\section{Conceptual view of intent-driven navigation}
\label{app:teaser-discussion}

\begin{figure}[h]
  \centering
  \includegraphics[width=\linewidth]{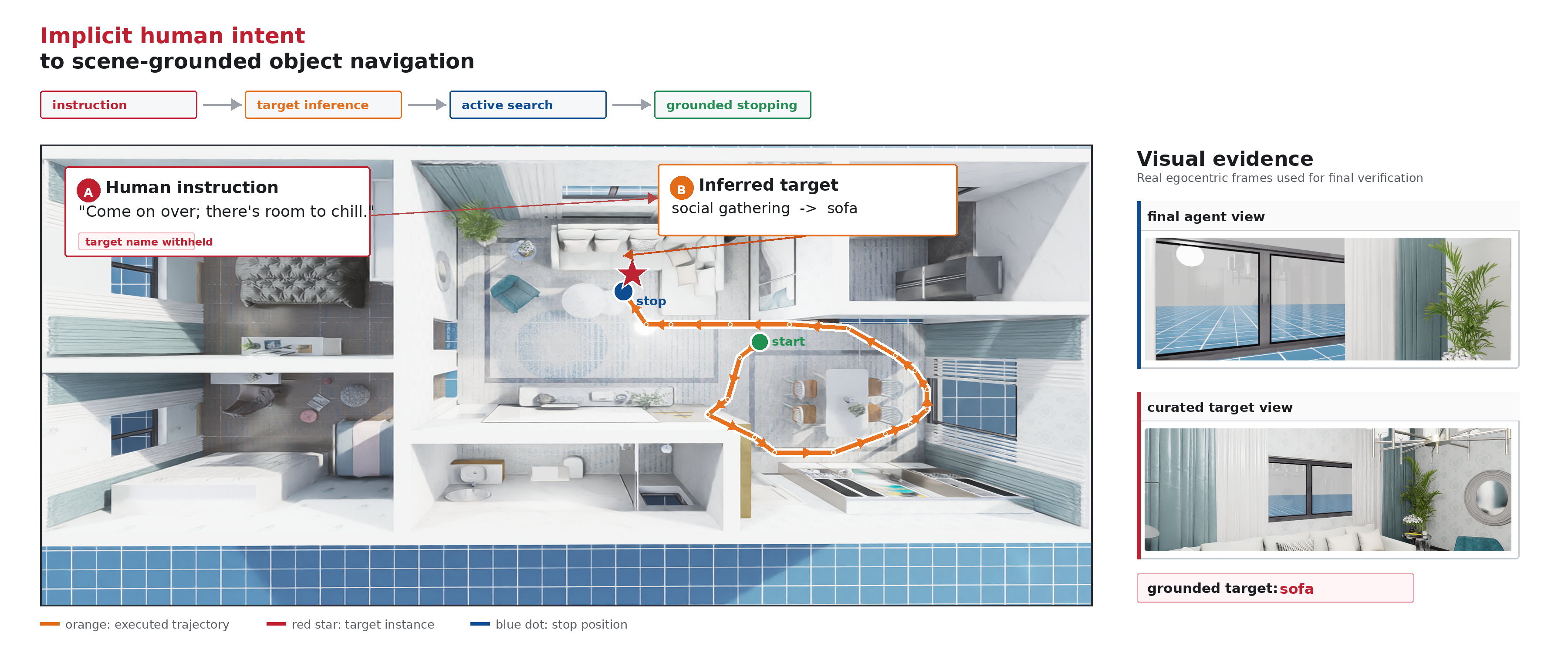}
  \caption{Expanded conceptual illustration of \ours{}. An implicit human request
  induces multiple plausible target hypotheses; active exploration gathers visual
  and spatial evidence; the agent succeeds only if it grounds the intent to the
  correct physical instance and terminates near that object. The diagram is
  explanatory rather than an additional quantitative result.}
  \label{fig:intentionnav-teaser-v2}
\end{figure}

Figure~\ref{fig:intentionnav-teaser-v2} provides an expanded conceptual view of
the task interface and the main inference bottleneck studied by \ours{}.
Unlike category-goal ObjectNav, where the target class is supplied as an input,
the agent begins from an underspecified human need.
The first stage therefore asks the model to map an intent expression to a set of
plausible object hypotheses rather than to a single named category.
For example, an instruction about warming food may initially support several
household objects; only some of these hypotheses are physically present, visible,
and reachable in the current scene.

The middle stages illustrate why this setting is not reducible to static visual
grounding.
The agent must actively collect evidence under partial observability: each RGB-D
view can confirm or reject candidate objects, while accumulated map memory links
local visual evidence to global scene position.
This separates three sources of error that are conflated in a single success
rate: semantic intent inference, visual evidence acquisition, and terminal
localization.
The final stage emphasizes that success is instance-grounded rather than
category-level: the agent must stop near the scene-grounded object that satisfies
the inferred intent, not merely predict a plausible object name.
The confidence values in the figure are illustrative and are used only to
visualize hypothesis ranking; all reported quantitative results are computed by
the main evaluation protocol.

\section{Synonym vocabulary and validation prompt}
\label{app:synonyms-prompts}

\metric{IM} uses a finite category vocabulary rather than exact string matching.
The official matcher lowercases and snake-cases the model's predicted target, then
accepts either the canonical category key or an alias in the submitted file
\texttt{eval/vocab/category\_synonyms.yaml}.
Table~\ref{tab:synonym-samples} lists representative entries.
The vocabulary covers common English aliases and spelling variants but is not
intended to resolve arbitrary paraphrases.

\begin{table}[h]
  \caption{Representative entries from the submitted synonym vocabulary used
  by \metric{IM}.}
  \label{tab:synonym-samples}
  \centering
  \small
  \begin{tabular}{ll}
    \toprule
    Canonical category & Accepted aliases \\
    \midrule
    \texttt{basin}      & washbasin, wash\_basin, sink, bathroom\_sink \\
    \texttt{fridge}     & refrigerator, icebox \\
    \texttt{closestool} & toilet, wc, commode, water\_closet \\
    \texttt{couch}      & sofa, settee \\
    \texttt{table\_lamp} & desk\_lamp, lamp \\
    \texttt{television} & tv, tele, flatscreen, smart\_tv \\
    \texttt{range\_hood} & extractor, kitchen\_hood, stove\_hood, vent\_hood \\
    \bottomrule
  \end{tabular}
\end{table}

The automated construction judge uses \texttt{gemini-3-flash-preview} to score each
candidate item on two dimensions: target grounding (whether the intent uniquely
implies the target given the scene photo) and style distinguishability (whether the
four variants are clearly distinct in register).
Items scoring below threshold on either dimension are revised for up to three rounds
or replaced.
The prompt template is embedded in \texttt{refine/refine\_unpolished.py} and is
distinct from the active-navigation agent prompts in \texttt{eval/judge/prompts/}.

\begin{quote}
\footnotesize
\begin{verbatim}
System:
You are a strict evaluator for a vision-language benchmark. You are
shown a room photo, a target object category, and four English
paraphrases of a user's intent. Your job is to decide how well those
four English variants, together with the photo, let a reader uniquely
identify the target object category. Return ONLY a single JSON object.

User:
Photo: (attached)

Target object category (the English variants should NOT name this
word directly): "{target_category}"

Difficulty label: {difficulty}

Four English style variants to evaluate:
  formal_en:    "{formal_en}"
  natural_en:   "{natural_en}"
  casual_en:    "{casual_en}"
  emotional_en: "{emotional_en}"

Rate 0-10 on EACH of these two dimensions. Integers only.

1. target_grounding (0-10)
   Given ONLY the attached photo + the target category word + these 4
   English variants, could a reader uniquely pick out the target object?
   10 = clearly and uniquely points to target; >=2 variants carry an
        affordance hint distinguishing this target from competitors.
    7 = target can be picked out, but one or two variants are generic.
    4 = generic; another visible object could equally satisfy it.
    0 = unrelated to target or points to a competitor.

2. style_distinguishability (0-10)
   Are the four variants clearly different in register/tone/length?
   formal_en:    polished, friendly-formal.
   natural_en:   everyday spoken English, contractions.
   casual_en:    brief (4-9 words), contractions.
   emotional_en: emotional/atmospheric, still conversational.
   10 = all four clearly distinct.
    7 = three distinct, one close to another.
    4 = two or more variants read very similarly.

Provide a short note (<=50 words): name the weakest variant and why,
or "all four look good".

Return JSON exactly:
{"target_grounding": <int 0-10>,
 "style_distinguishability": <int 0-10>,
 "notes": "<string>"}
\end{verbatim}
\end{quote}

\begin{figure}[t]
  \centering
  \small
  \begin{tikzpicture}
    \def\R{2.25}
    \foreach \r/\lab in {0.45/0.2,0.90/0.4,1.35/0.6,1.80/0.8,2.25/1.0} {
      \draw[gray!18] (90:\r) -- (30:\r) -- (-30:\r) -- (-90:\r) -- (-150:\r)
                     -- (150:\r) -- cycle;
      \node[gray!55,font=\scriptsize] at (83:\r) {\lab};
    }
    \draw[gray!35] (0,0) -- (90:\R);
    \node[font=\scriptsize,align=center,anchor=south] at (90:2.76) {Intent\\match};
    \draw[gray!35] (0,0) -- (30:\R);
    \node[font=\scriptsize,align=center,anchor=west] at (30:2.76) {2.0 m\\success};
    \draw[gray!35] (0,0) -- (-30:\R);
    \node[font=\scriptsize,align=center,anchor=west] at (-30:2.76) {1.0 m\\grounded};
    \draw[gray!35] (0,0) -- (-90:\R);
    \node[font=\scriptsize,align=center,anchor=north] at (-90:2.76) {Neighborhood\\reach};
    \draw[gray!35] (0,0) -- (-150:\R);
    \node[font=\scriptsize,align=center,anchor=east] at (-150:2.76) {Style\\stability};
    \draw[gray!35] (0,0) -- (150:\R);
    \node[font=\scriptsize,align=center,anchor=east] at (150:2.76) {Mode\\stability};
    \draw[blue!70!black,fill=blue!45,fill opacity=0.13,line width=0.8pt]
      (90:1.134) -- (30:0.578) -- (-30:0.149) -- (-90:1.551)
      -- (-150:2.174) -- (150:1.998) -- cycle;
    \draw[orange!85!black,fill=orange!55,fill opacity=0.11,line width=0.8pt,dashed]
      (90:1.134) -- (30:0.561) -- (-30:0.115) -- (-90:1.554)
      -- (-150:2.187) -- (150:1.993) -- cycle;
    \draw[green!55!black,fill=green!45,fill opacity=0.10,line width=0.8pt,
          densely dotted]
      (90:0.992) -- (30:0.543) -- (-30:0.108) -- (-90:1.533)
      -- (-150:2.178) -- (150:2.010) -- cycle;
    \begin{scope}[xshift=3.25cm,yshift=1.35cm]
      \draw[blue!70!black,line width=0.8pt] (0,0) -- (0.42,0);
      \node[anchor=west,font=\scriptsize] at (0.52,0) {Gemini-3.1};
      \draw[orange!85!black,line width=0.8pt,dashed] (0,-0.32) -- (0.42,-0.32);
      \node[anchor=west,font=\scriptsize] at (0.52,-0.32) {GPT-5.4};
      \draw[green!55!black,line width=0.8pt,densely dotted]
        (0,-0.64) -- (0.42,-0.64);
      \node[anchor=west,font=\scriptsize] at (0.52,-0.64) {Qwen3.6};
    \end{scope}
  \end{tikzpicture}
  \caption{Radar diagnostic profile of the three evaluated models. Axes are
  normalized to $[0,1]$; larger is better on all axes. Intent match, 2.0\,m
  success, 1.0\,m grounded success, and neighborhood reach correspond to the
  metrics in Table~\ref{tab:main-results}; style and mode stability are
  $1-\Delta_{\text{style}}$ and $1-\Delta_{\text{intent}}$ from
  Tables~\ref{tab:style} and~\ref{tab:intent-mode}.
  The three models cluster closely, confirming that the dominant signal is the
  shared gap structure rather than inter-model differences.}
  \label{fig:diagnostic-radar}
\end{figure}
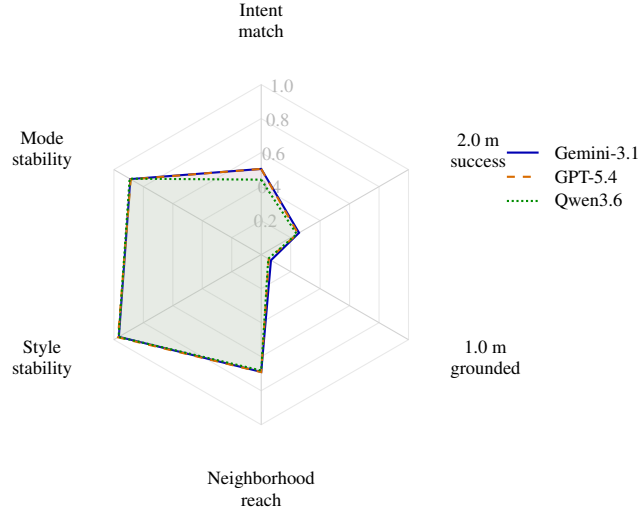

\section{Benchmark documentation and artifact scope}

\subsection{Motivation}
\ours{} is designed to measure how embodied agents ground indirect human intents
into active object navigation.
The evaluation split is intended for benchmark evaluation and diagnostic analysis, not for
training deployed systems.
The primary users are researchers working on VLM-guided navigation, open-vocabulary
perception, and human-robot interaction.

\subsection{Composition}
Each item contains a scene identifier, target category, target object identifier,
target position, original intent scaffold, four English instruction styles, and
intent-mode metadata.
The benchmark contains 500 unique intents, 2000 English instructions, 176 scenes,
64 target categories, and four primary intent modes.
The benchmark records do not contain personal information, human-subject records, or real
household user logs.

\subsection{Collection and preprocessing}
Scene objects are captured from VLNVerse Isaac Sim environments \citep{lin2025vlnverse}.
Intent candidates are generated from scene-target visual context, then rewritten into
four English styles under constraints that prevent direct target naming.
A VLM judge scores target grounding and style distinguishability; low-scoring items
are refined iteratively.
The final subset is selected for target diversity and physical uniqueness, so that
each intent has a single scene-grounded target for evaluation.

\subsection{Recommended use}
Researchers should report \metric{IM}, \metric{SR}, \metric{GSR}, \metric{OSR},
\metric{SPL}, and \metric{TL}, together with style- and mode-stratified results.
\metric{SR}, \metric{OSR}, and \metric{SPL} use the 2.0\,m headline radius;
\metric{GSR} uses 1.0\,m with final-frame target visibility.
The benchmark is most informative for comparing failure modes across the
intent-to-navigation chain rather than ranking model families.

\subsection{Artifact scope and maintenance}
The submitted code artifact includes the metric implementation, prompt and
configuration files, and reproducibility tools that regenerate tables when supplied
with compatible raw episode logs.
The 500-intent split used for reported results is fixed; future larger splits
should receive a new version identifier.
Bug fixes are documented as either metadata-only corrections or
evaluation-affecting changes.
If simulator assets or external model services become unavailable in future uses,
compatible replacements should be documented and resulting numbers should be marked
as a new benchmark version.

\subsection{Licenses and access}
The benchmark depends on the licenses of the simulator, scene assets,
Grounding-DINO, VLM providers, and generated annotations.
Generated benchmark annotations and full scene-derived assets are not included in
the submitted code artifact.
Service-based evaluations record model names, provider dates, and decoding settings
because commercial model behavior may change over time.

\subsection{Ethical considerations}
The benchmark is simulated and does not include human subjects, but its intended
application is human-facing embodied assistance.
Agents that infer intent may make unsafe assumptions when requests are ambiguous,
culturally specific, or underspecified; results on \ours{} should be interpreted as
diagnostic evidence, not proof of safe real-world deployment.
Future extensions should include multilingual intents, dialogue-based clarification,
and fairness analysis over diverse user expression styles.

\end{document}